\documentclass[letterpaper, 10pt, twocolumn]{article}
\usepackage{clean}
\shorttitle{Compliant gripper for assembly of electrical connectors}

\usepackage{gensymb}
\usepackage{adjustbox}
\usepackage{graphicx}
\usepackage{subcaption}
\graphicspath{{./figs}}
\usepackage{epsfig}

\usepackage{amsmath,amssymb,amsfonts}
\usepackage{algorithm2e}
\usepackage{xcolor}
\usepackage[colorlinks=true, linkcolor=black, urlcolor=cyan, filecolor=black, citecolor=black]{hyperref}
\usepackage{url}
\usepackage{amsmath,bm}
\usepackage{multirow, makecell} 
\usepackage{adjustbox}
\usepackage{comment}

\usepackage{xcolor}   
\newcommand{\rev}[1]{{\color{black} #1}}  
\begin{document}

\title{High-speed electrical connector assembly by structured \\ compliance in a finray-effect gripper}
\author{Richard Matthias Hartisch$^1$ Kevin Haninger$^2$ 
\thanks{\noindent$^1$ Department of Industrial Automation Technology at TU Berlin, Germany. \\ $^2$ Department of Automation at Fraunhofer IPK, Berlin, Germany.  \\ Corresponding author: {\tt r.hartisch@tu-berlin.de}}
\thanks{This project has received funding from the European Union's Horizon 2020 research and innovation programme under grant agreement No 820689 — SHERLOCK and 101058521 — CONVERGING.}}

\maketitle

\begin{abstract}

Fine assembly tasks such as electrical connector insertion have tight tolerances and sensitive components, \rev{requiring compensation of alignment errors while applying sufficient force in the insertion direction, ideally at high speeds and while grasping a range of components.} Vision, tactile, or force sensors can compensate alignment errors, but have limited bandwidth, limiting the safe assembly speed. \rev{Passive compliance such as silicone-based fingers can reduce collision forces and grasp a range of components, but often cannot provide the accuracy or assembly forces required. To support high-speed mechanical search and self-aligning insertion, this paper proposes monolithic additively manufactured fingers which realize a moderate, structured compliance directly proximal to the gripped object. The geometry of finray-effect fingers are adapted to add form-closure features and realize a directionally-dependent stiffness at the fingertip, with a high stiffness to apply insertion forces and lower transverse stiffness to support alignment. Design parameters and mechanical properties of the fingers are investigated with FEM and empirical studies, analyzing the stiffness, maximum load, and viscoelastic effects. The fingers realize a remote center of compliance, which is shown to depend on the rib angle, and a directional stiffness ratio of $14-36$. The fingers are applied to a plug insertion task, realizing a tolerance window of $7.5$ mm and approach speeds of $1.3$ m/s.}
\end{abstract}

\section{Introduction}
\label{sec:intro}
The installation of cables and wire harnesses has substantial industrial demand, especially as electrification of automobiles and household appliances increases. While the pre-production of cable harnesses (cutting, mounting of wire seals and attachment of cable heads) can be achieved with specialized machinery \cite{trommnau2019overviewneu}, installation is today largely manual work \cite{Yumbla.2020}. 

Cable installation is challenging to automate due to the high variety in connectors \cite{Yumbla.2020} which can lead to small batch sizes \cite{trommnau2019overviewneu}. The manipulation of cables also introduces technical challenges, where cable routing requires perception and planning methods for deformable linear objects \cite{trommnau2019overviewneu, Chen.2016}. However, connector insertion is a critical functionality for all cable installation tasks. It is also challenging: it requires coordinated vision and touch when done by humans \cite{Chen.2016}, and robotic solutions often require a combination of sensors (vision \cite{trommnau2019overviewneu}, tactile \cite{li2014localization,   bhirangi2021reskin, she2021cable}, or force \cite{ Haraguchi.2011}) and application-specific finger design \cite{chapman2021locally, chen2012design, Yumbla.2019}.

\begin{figure}[t]
  \begin{center}
  \begin{subfigure}[!]{0.65\linewidth}

      \includegraphics[width=\columnwidth]{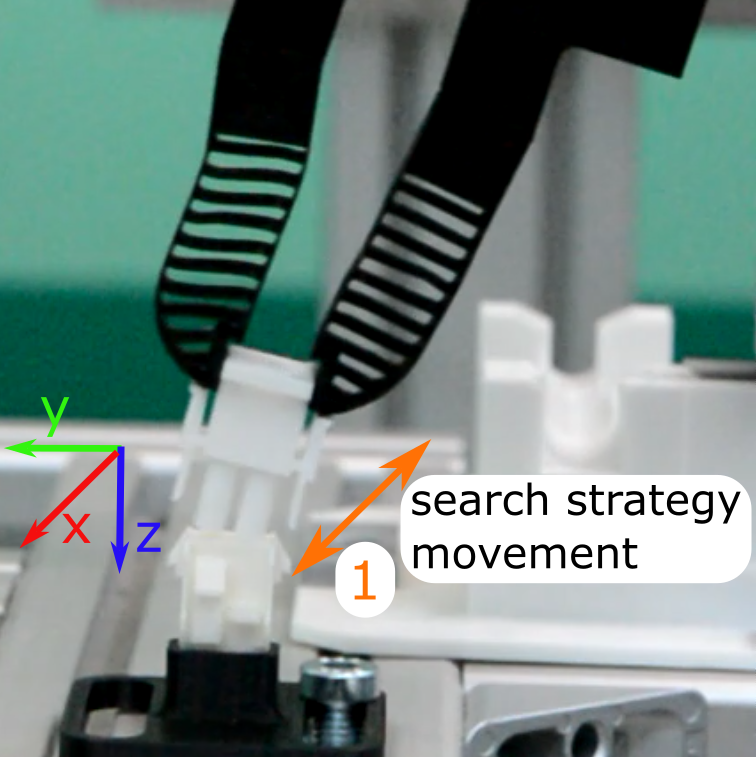}
    \caption{}\label{subfig:key-a}

  \end{subfigure}
\end{center}
  \hfill
  \vfill
  \begin{minipage}{0.495\linewidth}
    \begin{subfigure}{\linewidth}
      \includegraphics[width=\linewidth]{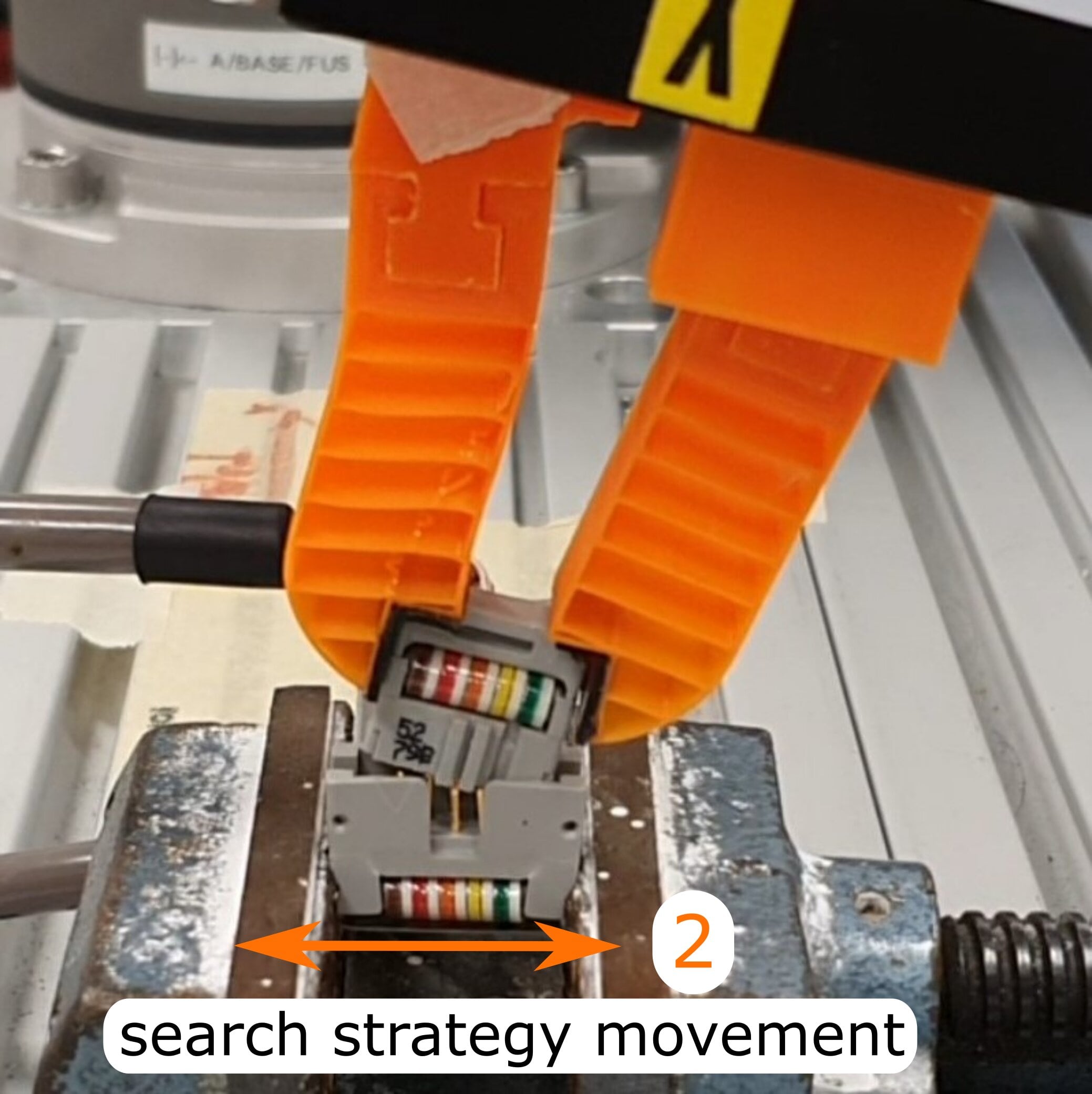}
      \caption{}\label{subfig:key-b}
    \end{subfigure}
      \end{minipage}
    \hfill
      \begin{minipage}{0.495\linewidth}
    \begin{subfigure}{\linewidth}
      \includegraphics[width=\linewidth]{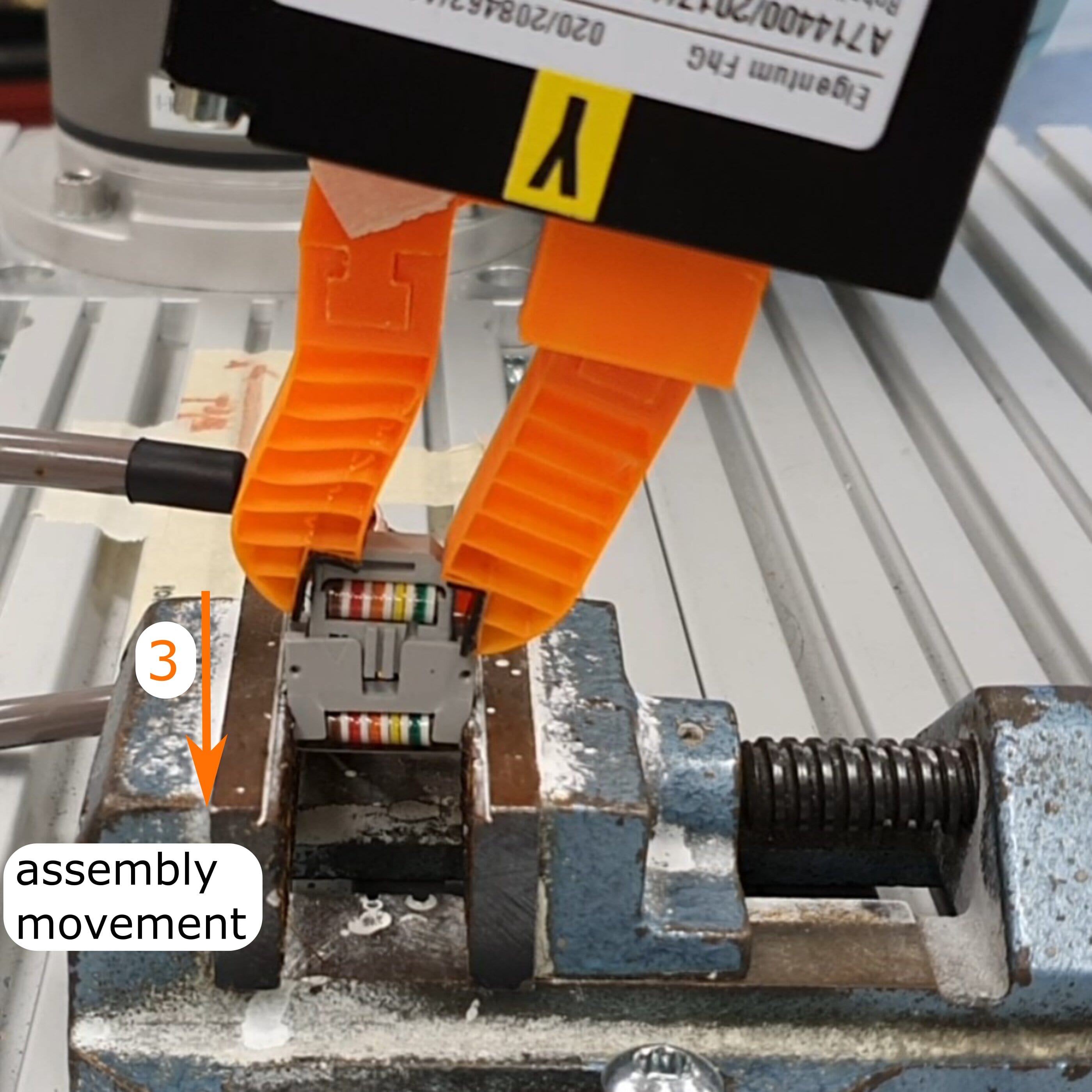}
      \caption{}\label{subfig:key-c}
    \end{subfigure}\hfill
  \end{minipage}
  \caption{The proposed fingers are demonstrated with varying types of plugs, showing the working principle and search strategy, where (a) shows the first movement of the search strategy and the coordinate system, (b) and (c) visualize the final search and assembly movements.}
  \vspace{-0.7cm}
  \label{fig:working principle}
\end{figure}

\rev{The challenges in the gripping, alignment, and insertion of electrical connectors are largely shared by other fine assembly tasks. In both cases, alignment errors can cause failed assembly \cite{Yumbla.2019}, requiring the use of some error compensation methods \cite{li2019c}. Assembly force must be applied to bring together a snap or friction fit without causing jamming \cite{li2019c}. Inserting in obstructed and complex environments often imposes tight space requirements on the gripper, and the need to leave the inserted portion free often requires a fingertip grip \cite{whitney2004mechanicalneu}.}

These challenges can be partly handled by compliance. In insertion, compliance can compensate misalignment between gripped part and socket \cite{whitney2004mechanicalneu, li2019c}.  Compliance can be active or passive \cite{wang1998passive, li2019c}, where passive compliance is the intrinsic mechanical compliance of the physical structure, and active compliance is achieved by feedback controller design. A relevant example for compliance is the remote center of compliance (RCC) \cite{ciblak2003designneu, whitney2004mechanicalneu}, which allows self-alignment in insertion tasks \cite{yun2008a}. Active compliance, such as impedance or admittance control can adapt the relative pose according to contact forces \cite{baksys2017vibratory, li2019c}. 

The major advantage of active compliance is the possibility to digitally change compliance, e.g. adjusting the RCC location to improve performance \cite{wang1998passive} or using general non-diagonal stiffness matrices \cite{oikawa2021}. The disadvantages of active compliance are the relatively high costs and the limited bandwidth \cite{wang1998passive}, which typically leads to higher collision forces. In contrast, passive compliance has \rev{almost} no bandwidth limits, allowing high-speed contact transitions \cite{bicchi2004}. \rev{However, many materials used in soft robotics have viscoelastic effects \cite{shintake2018}, which can affect high-speed performance.} Physical compliance also reduces the positioning accuracy and ability to detect contact \cite{li2019c, haninger2022b}, but supports motion strategies which exploit contact such as mechanical search \cite{hamaya2020} with self-alignment \cite{Yun.2008}. 

\rev{Passive compliance can be integrated at various locations in the robot; the joints \cite{albu-schaffer2008}, flange \cite{whitney2004mechanicalneu}, fingers \cite{shintake2018} or environment \cite{hartisch2022flexure}. By integrating the compliance directly proximal to the contact, the sprung inertia is reduced, reducing inertial forces at high accelerations. The remote center of compliance can be integrated in the robot flange, but then gripper contributes to sprung inertia, typically contributing $>1$ kg of inertia. On the other hand, soft fingers can provide proximal compliance, as well as realizing a larger contact area over variation in the gripped object's surface geometry \cite{shintake2018}.}

\rev{Soft fingers can be realized with silicone \cite{hao2016universal, manti2015bioinspired, park2018hybrid, liu2020two}, often with the objective of universal gripping. However, these silicon-based soft fingers often render a very low stiffness \cite{hernandez2023}, making it difficult to realize the repeatability or assembly forces needed in many fine assembly tasks \cite{li2019c}.  Soft fingers can be constructed by additive manufacturing, which can be used to construct monolithic soft pneumatic fingers \cite{Tawk.2019} or flexure-based grippers \cite{hernandez2023}.}

\rev{Designing passive compliance for assembly is typically done in terms of the spatial stiffness rendered on the gripped object, which affects collision force \cite{bicchi2004} and alignment error compensation \cite{Yun.2008, huang1998}. Modelling of soft fingers is often done with continuum mechanics, Cosserat rod theory or FEM \cite{schegg2022}, which can support kinematic design and control, but often does not consider rendered stiffness. The parallel or series combination of elemental stiffness can be used to analyze rendered stiffness \cite{huang1998, hartisch2022flexure}, but this is difficult to scale to complex soft fingers.} The limited \rev{a priori design methods for} rendered directional compliance leads to a need for iterative testing for a specific task or part \cite{chen2017improved}. However, this iterative testing can be accelerated with fast prototyping methods like fused deposition modelling \cite{hartisch2022flexure, Elgeneidy.2019}, which can also integrate structural passive compliance in monolithic structures with flexures \cite{hernandez2023}. 

This work proposes structured compliant fingers for electrical component assembly, using fused deposition modeling to produce low-cost monolithic devices which can be \rev{easily integrated to parallel grippers}. \rev{Compared with soft fingers which target universal gripping, either silicone-based \cite{hao2016universal, manti2015bioinspired, park2018hybrid, liu2020two} or finray-based \cite{Elgeneidy.2019}, these fingers realize a structured compliance for passive alignment while providing sufficient force in the assembly direction.} Compared with sensorized fingers for plug insertion \cite{wang2021, jiang2022}, the proposed compliance allows a larger tolerance window and higher speed. \rev{In contrast to} plug insertion approaches using active compliance, which can take up to $\approx 6-16$s from first contact to insertion \cite{park2013intuitive}, the passive compliance allows a successful assembly of the connectors in $\approx 1.2$s. This paper also provides a taxonomy and detailed requirements of the connector insertion problem, whereas existing work \cite{chen2012design, Yumbla.2019, trommnau2019overviewneu} focus on wire harness design and production, not connector mating. 

A previous version of this paper was submitted to the 2023 IEEE/ASME International Conference on Advanced Intelligent Mechatronics \cite{hartisch2023compliant}. This paper adds \rev{analysis of the connector assembly problem, FEM and emprical identification of the directional finger stiffness, empirical investigation of viscoelastic effect,} as well as assembly applications with higher speeds \rev{and a wider range of components}. 

The paper is organized as follows. Section \ref{sec:problem_desc} categorizes the parameters of the plugs used in this work, parameters occurring in the assembly task, and describes the steps of the assembly process. Section \ref{sec:design} introduces the final gripper design, derived from the finray-effect, where the design parameters, the design and manufacturing process, and problems are described. A range of applications to verify the gripper's abilities are presented in Section \ref{sec:app}, consisting of repeatability and robustness experiments to determine design parameters which achieve the widest tolerable scale of misalignment. Finally, the conclusion and future work is given in Section \ref{sec:discussion}.

\section{Electrical connector problem description}
\label{sec:problem_desc}

This section analyzes the problem of connector assembly, providing a taxonomy of electrical connectors and the assembly process itself.

\subsection{Taxonomy of connectors \label{sec:taxonomy}}
\begin{table*}[ht]
\renewcommand{\arraystretch}{1.15} 
\begin{center}
 \vspace*{.2cm}
\caption{Identified important properties of connectors and the assembly task, what parts of the robotic solution they influence, and possible values that the property can take \label{tab:conn_parameters}}
\begin{tabular}{r r|l|l} 
& Property & Effects & Possible values \\
\hline
\multirow{5}{*}{\rotatebox[origin=c]{90}{\large Connector}} & Fit and tolerances & Search strategy, req'd assembly force & Press, running, transition \\
& Plug exposed after insert & Grip location in insertion & Flush, $>0$ mm \\
 & Cable gland orientation & Grip location and free space & Straight, right angle \\
& Pin height & Search strategy & Flush, $<0$ mm \\
& Locking feature & Insertion, validation & Clip, lever  \\
\hline
\multirow{5}{*}{\rotatebox[origin=c]{90}{\large Task}} & Plug availability & Grasp strategy, finger design & magazine, on table, cluttered \\
& Socket availability & Tolerances, search strategy & fixed position, in workpiece, free space \\
& Space requirements & Finger dimensions, robot strategy & free space dimensions \\
& Cable handling & Finger design, expected force profile & need insert clips, need to pull cable \\
& Validation & Insert strategy & Is a validation (e.g. push-pull-push) required? \\
\hline

\end{tabular}
\vspace*{-0.5cm}
\end{center}
\end{table*}
While there is large variation in connector design  \cite{Yumbla.2020}, the parameters summarized in Table \ref{tab:conn_parameters} have a significant influence on the finger design and strategies for grasping, searching, or insertion. 

Some parameters are shown visually in Figure \ref{fig:plug_grip}, left, which shows an inserted plug. The amount that the cable head extends from the socket after insertion limits the feasible gripping area. The cable gland can have different orientations, either straight out or in a right angle into the plug, which changes what space must be left free by the finger design. The cable type can be categorized as either a single-, ribbon-, or multi-cable.  Furthermore, the pin height inside the plug and/or socket can result in collision and jamming for some search strategies. Additional locking features, such as levers or clips, may require additional assembly force or post-processing to secure. The tolerances between the plug and socket influence the search strategy and required assembly force.

\subsection{Categorization of assembly task}

There are additional parameters in typical connector assembly tasks which affect the design. How the plug is supplied affects the uncertainty in grip pose, as the plug could either be fixed rigidly, e.g. in a magazine, lying freely on the table or placed in a cluttered environment. Similarly, the socket could be either be in a fixed position, integrated in a workpiece or in a magazine. Further, space limits from the environment can limit both the finger dimensions and robot search strategy. During assembly, additional cable and wire handling has to be considered, e.g. intermediate clips or cable straightening. After successfully mating the components additional testing could be necessary, e.g. push-pull-push of the cable. 

\subsection{Grip, search, and insert strategies}
\label{subsec:strategies}
The complete assembly process is considered in three stages: grip, search, and insert. From an initial position, the plug is gripped. Without a magazine or jig providing a constant and known pose of the plug, a known pose or at least the orientation of the plug inside the grip should be established within the tolerances of the insertion process. 
\begin{figure}[t]
\begin{center}
\includegraphics[width=0.38\textwidth]{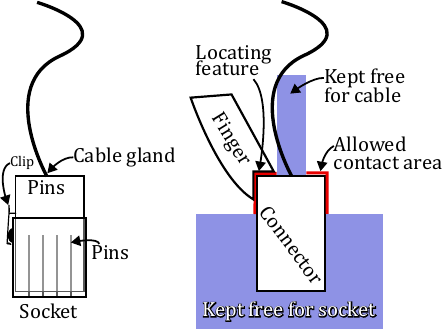}
\end{center}
\caption{Left, inserted connector in socket with key features, right, finger grasping the connector \label{fig:plug_grip}}
\vspace{-0.5cm}
\end{figure}

The grasping strategy includes aspects of the finger design seen in the right of Fig. \ref{fig:plug_grip}: (i) what contact area between the plug and finger can be used. Multiple grasping strategies are possible here, either a pinch contact or parallel grasp of the wires or the cable head, (ii) what space around the connector must be kept free, (iii) are locating features needed to provide either repeatable position or sufficient assembly force? In addition to the fingertip design, the grasping strategy may include (i) a magazine for providing the plug in a semi-repeatable way and (ii) an adjustment strategy to ensure the connector is in a repeatable position in the fingers.  

The search strategy should achieve alignment of the plug and socket. When using mechanical search, the search strategy should be designed considering: (i) the variation in pose that needs to be covered with the strategy, (ii) the initial contact between plug and socket, which can be a point, line or planar contact depending on how the plug is presented, (iii) the height of the pins, which could be bent if contacted by the tip of the plug, (iv) validating that the plug has successfully slipped into the socket after the alignment.  

The mechanical search strategy used here is shown in Fig. \ref{fig:working principle}.  Initial contact is made with a tilted plug, such that the corner of the plug lightly presses on the edge of the connector, Fig. \ref{subfig:key-a}. A motion in the x-direction allows the plug to slip into the socket when aligned. Next, contact is established with the sides of plug and socket, realized by a motion in the y-direction, Fig. \ref{subfig:key-b}. At this point, the leading corner of the plug should be slightly inserted and resting on the edge of the socket. \rev{The search strategy used in this work utilizes an open-loop position control, without the use of force feedback.}

For the insertion phase following aspects should be considered: (i) the finger design should be able to avoid jamming of the connectors. For this, compliance, either active or passive, could be suitable, where the plug is able to rotate inside the socket due to the contact. (ii) The assembly force should not exceed a certain threshold, to avoid damaging the parts, which can also be realized by the finger compliance.

\section{Design of Compliant Finray-effect Grippers}
\label{sec:design}

In this section, we describe the modification and parameterization of a finray-effect gripper \cite{crooks2016fin} to \rev{meet the requirements of Section \ref{sec:taxonomy}}. Where classical finray-effect grippers \rev{use an enveloping grip where} deformation adapts to variation in surface geometry of grasped parts \cite{Elgeneidy.2019}, we propose a design where the \rev{components are grasped at the fingertip and} rendered compliance on a gripped part \rev{supports mechanical search, passive alignment, while providing sufficient force in the assembly direction}. 

Considering the requirements and constraints in Sec. \ref{sec:problem_desc}, the finray design is adapted with the flexibility of 3D printing to optimize the following design parameters, seen in Figure \ref{finger_design_rotated}: (i) fingertip design, (ii) rib angle / infill direction, (iii) rib density / infill density, (iv) finger mounting angle. 

\subsection{Modified Finray Design}

The finray-effect gripper mimics the deformation of fish fins, which are composed by two outer walls forming a V shape. Between the bones, crossbeams are placed which determine the mechanical properties of the finray-effect gripper. 
The side walls of the standard finray-effect gripper bend from contact when grasping convex parts, which results in a deformation of the base and tip towards the applied force \cite{crooks2016fin}. However, the standard V-shaped finray design is not ideal for the requirements here. Here, an object should be grasped at the fingertip, capable of applying high forces in the assembly direction and offering a lower stiffness laterally to compensate misalignment. 

When a flat object is grasped with a V-shaped fingertip, the assembly forces are carried only by friction, which may be insufficient for small components with small allowed grasping contact areas. A form-closure fingertip can therefore support higher assembly forces. Furthermore, to compensate misalignment for parts grasped at the fingertip: a V-shaped fingertip is stiff in the horizontal direction, and when the finger deforms, rotation of the fingertip occurs, which could result in contact loss with the gripped part. Here, a translational deflection at the fingertip is desired to compensate misalignment while maintaining the contact between gripper and grasped part. 

\rev{The V-shape finger profile is changed, instead of the outer walls approaching another towards the tip, the walls are more rectangular. With the fins, this produces a parallelogram structure which does not rotate the tip as it is displaced. Additionally, form-closure features are added at the fingertips, a shoulder which is discussed in the following Section \ref{subsec:design}. Due to the design of the notched fingertip, the generalizability to handle various connectors is limited, but the ability to apply forces over small contact areas is improved.}

\rev{With this fingertip, a directionally-dependent stiffness can be achieved, that is the effective stiffness between gripped object and robot is higher in the assembly direction, and lower in the orthogonal directions to support mechanical search and alignment compensation.}

\subsection{Design Parameters}  \label{subsec:design}
Two important parameters of the finger design are optimized to improve performance. 

\subsubsection{Infill Options}
The most important design parameters are the infill options to adjust the density and orientation of the ribs in the finger, i.e. the infill direction, given in $\degree$, and the infill density options, given in $ \%$, as proposed by \cite{Elgeneidy.2019} and visualized in Fig. \ref{finger_design_rotated}. This affects the bulk stiffness realized by the finger on a gripped part, \rev{effective remote center of rotation,} as well as the maximum force that the finger can apply\rev{, as shown in Section \ref{sec:single_finger_stiffness}}.

\subsubsection{Fingertip Options}
\label{subsub:fingertip}
An additional parameter is the form of the fingers which can either be with a rounded top, flat top, notched rounded top, flat angled or notched top with a contact plane, visualized in Fig. \ref{finger_design_rotated}. The notch can be rotated by a certain degree, corresponding to the mounting angle of the finger, 
used to achieve a parallel contact plane with the grasped part. The size of the notch depends on the connector to be handled, 
an oversized notch won't contact the plug or will interfere with insertion,
and an undersized notch has a smaller contact surface and could result in an unstable grip. 

\rev{While adding a notched fingertip results in form-fit contact, this limits the generalizability of connectors to be handled, which is why both characteristics have to be considered during the design phase. A notched fingertip could result in an ideal solution for one connector type, but could result in an unstable grip for other connectors.}

The contact surface also needs consideration: the friction of PLA+ and PETG are low, which is why additionally a thin rubber layer is applied after printing, which can be seen in Fig. \ref{fig:robot_zoom}. \rev{It is important to note, that the rubber layer consists of one adhesive layer to adhere on the contact plane of the finger. The side of the rubber tape in contact with the grasped part doesn't have an additional adhesive applied to it, only increasing the friction coefficient.}

\subsection{Manufacturing Process}
To allow for an easy adaptation of infill density and line directions, these parameters are set directly in the slicer program instead of CAD. Here \textit{Ultimaker Cura} is used, applying a method similar to the method used in \cite{Elgeneidy.2019}. The materials used in this work are orange PLA+  
and black PETG. 
\rev{ Other materials, such as TPU and ABS were found to be unsuitable. Tests indicated TPU's inherent compliance interferes with the effects of structured compliance, e.g. the material would be too compliant to provide the necessary stiffness in assembly direction. ABS has a tendency to warp during printing, which proved to be a major issue with the thin walls of this finger.}

\begin{figure}[t]
  \begin{subfigure}[!]{0.58\linewidth}
    \includegraphics[width=\columnwidth]{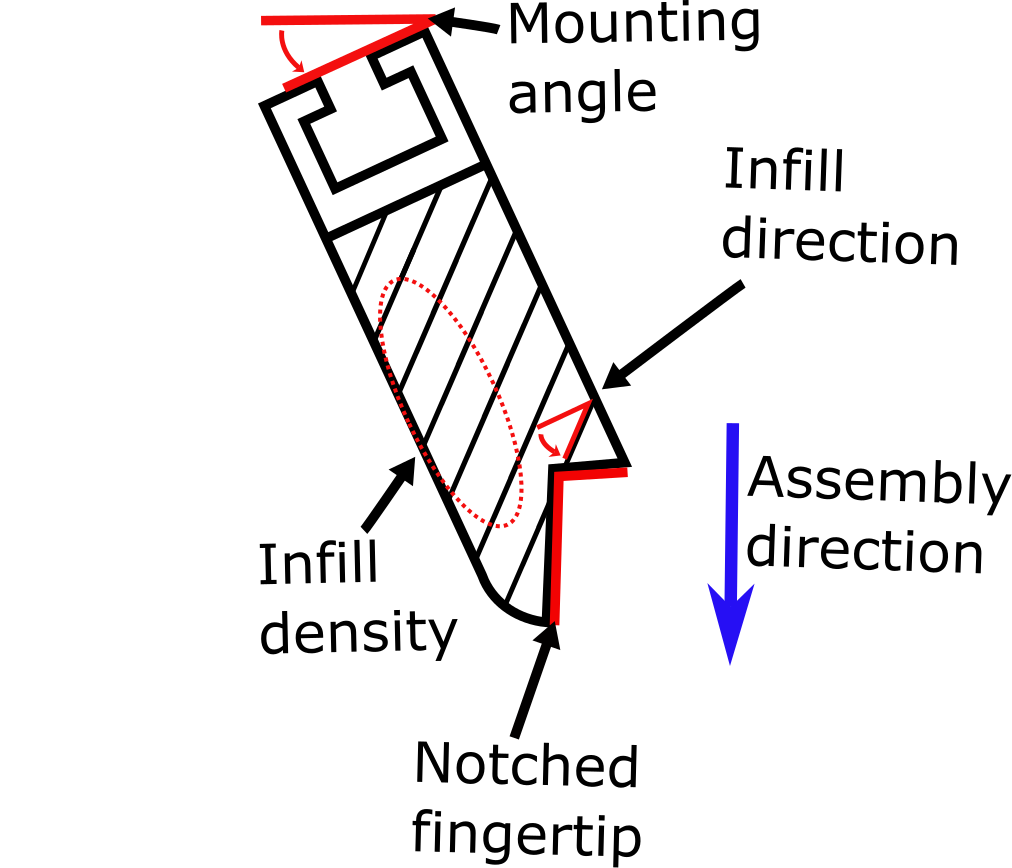}
    \caption{}\label{finger_design_rotated}
  \end{subfigure}\hfill
  \begin{minipage}{0.4\linewidth}
    \begin{subfigure}{\linewidth}
      \includegraphics[width=\linewidth]{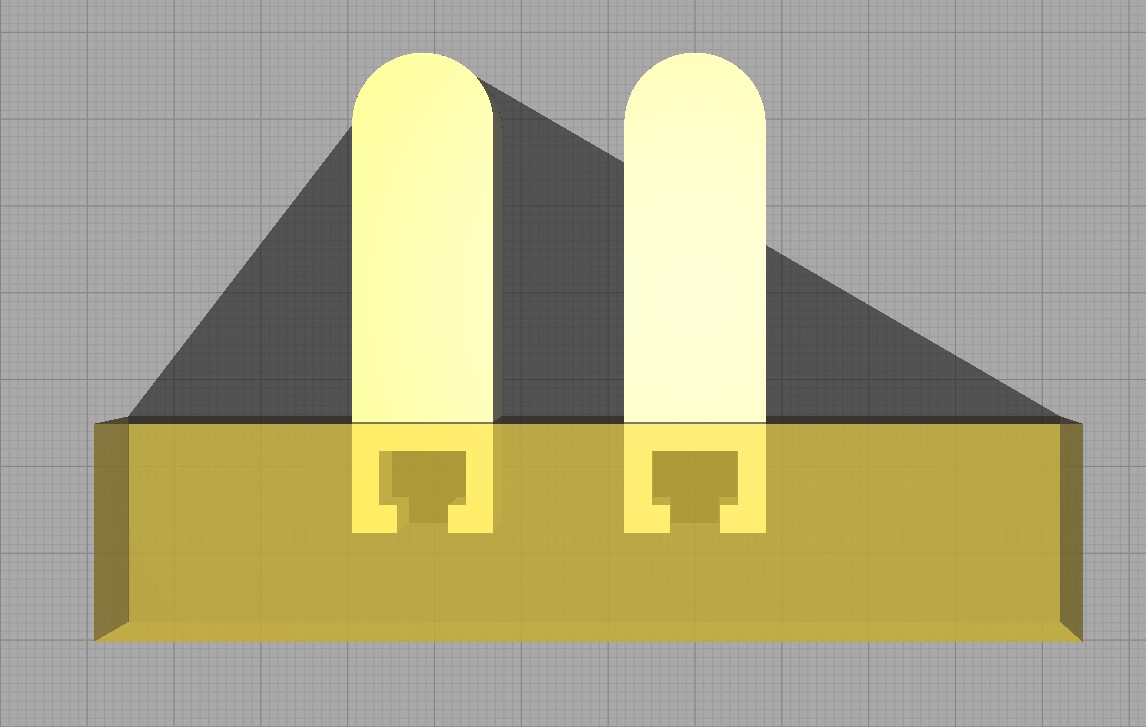}
      \caption{}\label{support_blocker}
    \end{subfigure}\hfill
    \medskip
    \begin{subfigure}{\linewidth}
      \includegraphics[width=\linewidth]{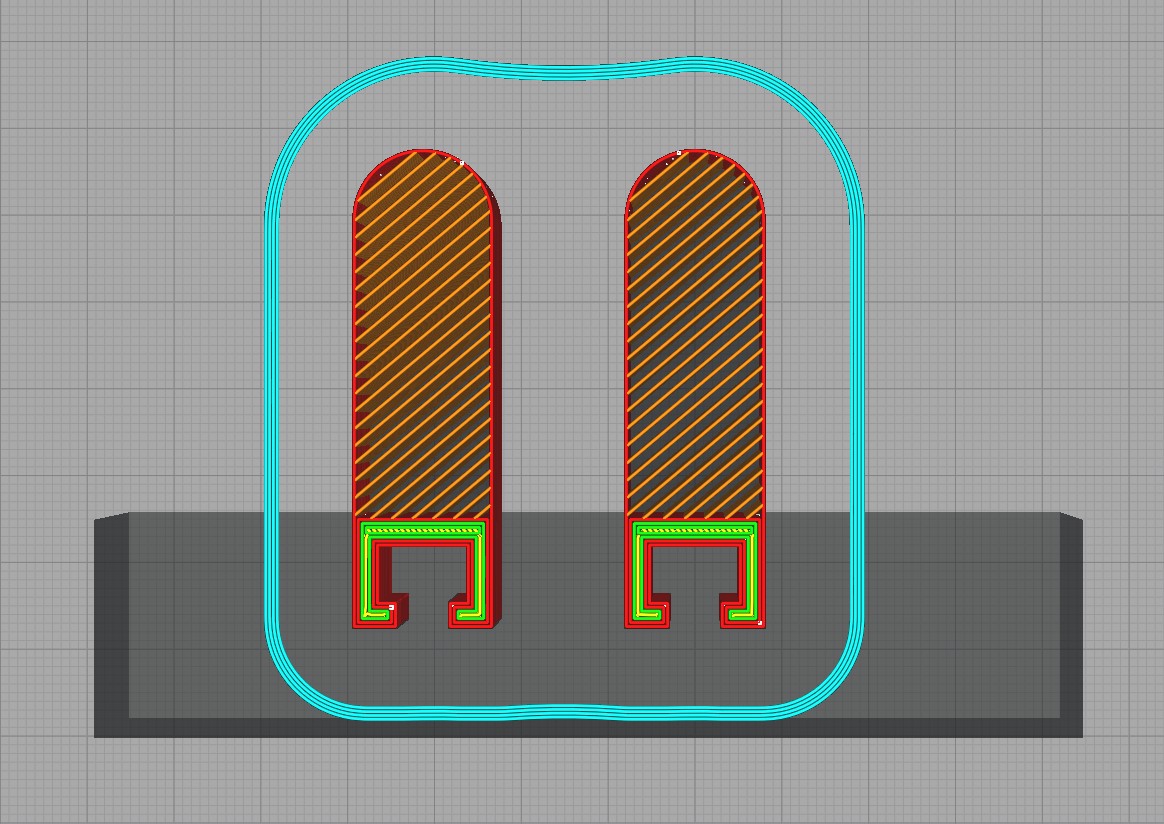}
      \caption{}\label{sliced}
    \end{subfigure}\hfill
  \end{minipage}
  \caption{(a) demonstrates the design parameters and the assembly directions. As shown here, the infill direction has an inclination of $40 \degree$. $0 \degree$ infill direction would result in ribs aligned parallel to the mounting surface. (b) and (c) show the design process of the finray-effect gripper: in (b) support blockers are demonstrated to allow for varying slicing settings, and (c) shows the sliced gripper with compliant structures at the top and a rigid base}
  \vspace{-0.5cm}
  \label{fig:design_process}
\end{figure}

\rev{
A CAD model of the solid finger is sliced  in \textit{Cura}, the infill type set to unconnected lines of width $0.4$ mm, the wall line count set to one with a line width close to the nozzle diameter of $0.4$ mm (less than the recommended $2\times$ nozzle diameter of \cite{Elgeneidy.2019}). To print the finger mount as a solid, \textit{Cura's} ``support blocker" feature is applied, as visualized in Fig \ref{support_blocker}. The .stl/.stp/.ipt files and further details on the manufacturing are available at \url{https://github.com/richardhartisch/compliantfinray}.}


\section{Mechanical Attributes}

\begin{figure} [t]
    \centering
    \subfloat[
	\label{fig:robot_left}]{\includegraphics[width=.4\columnwidth]{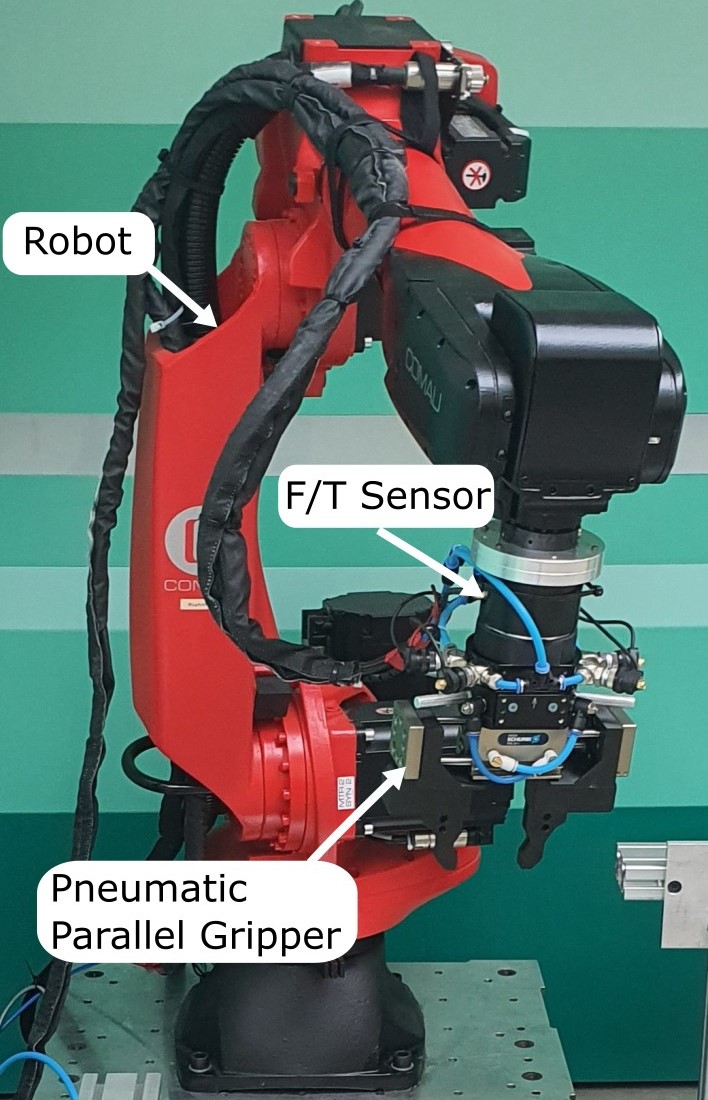}}
	\hfill
    \subfloat[
	\label{fig:robot_zoom}]{\includegraphics[width=.58\columnwidth]{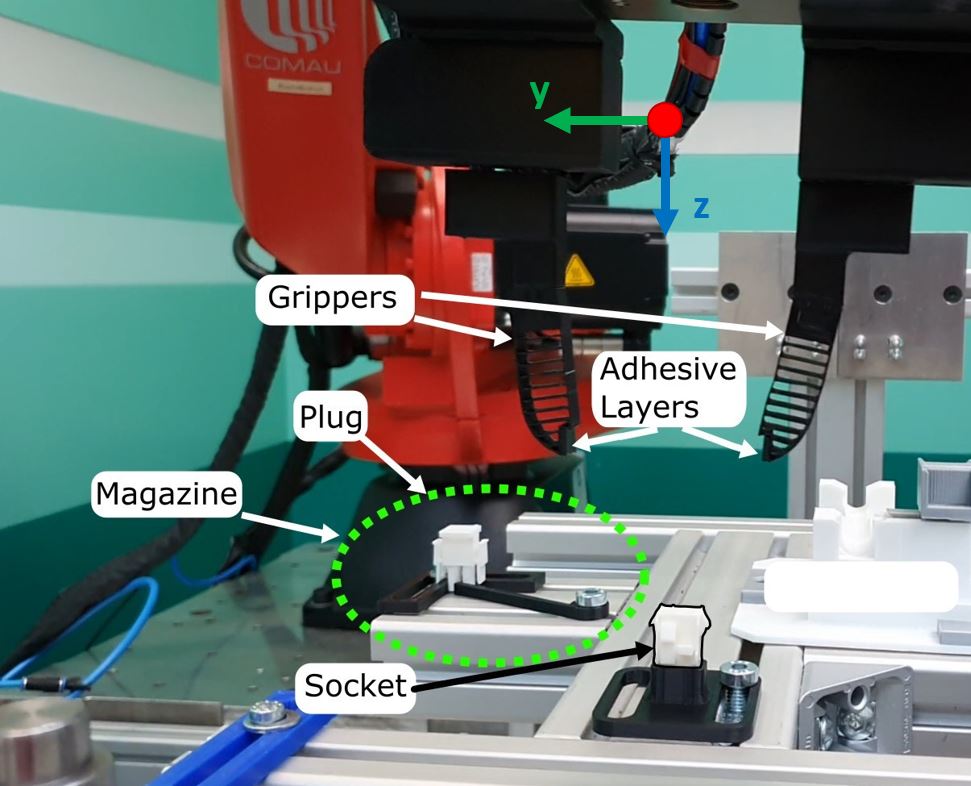}}
    \caption{In (a) and (b) the setup of the robot to determine the robustness and repeatability of the gripper is shown}
    \vspace{-0.5cm}
    \label{fig:exp_setup}
\end{figure}

\begin{figure}[t]
\begin{center}
\includegraphics[width=0.5\textwidth]{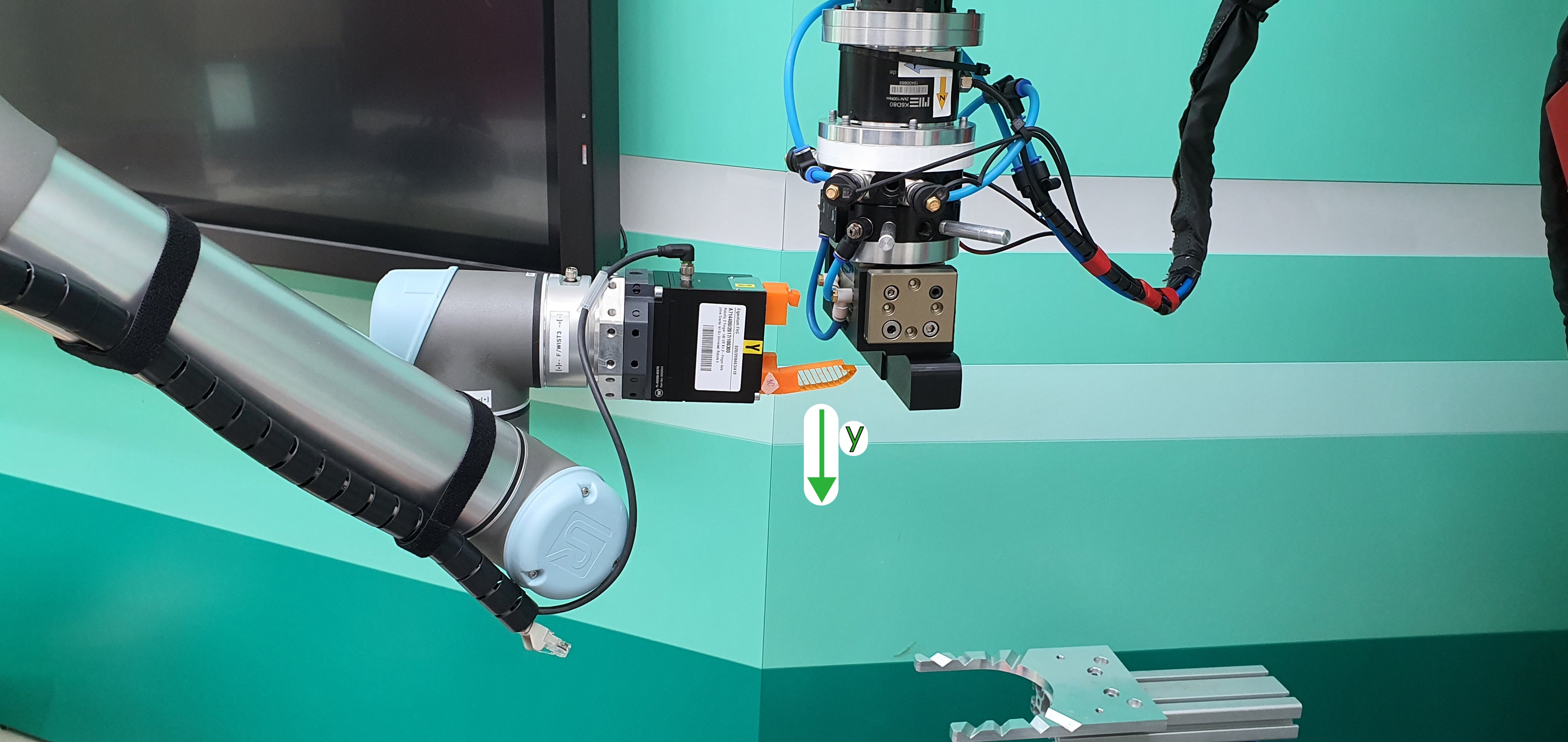}
\end{center}
\caption{Here, the setup of the robot to determine the mechanical properties of the gripper and how the gripper is being loaded is shown}
\vspace{-0.5cm}
\label{fig:stiffness_setup}
\end{figure}

In the following section, the mechanical properties of the finray-effect gripper are investigated: the stiffness of a single finger, viscoelastic effects, the maximum force and deflection until component failure is reached. The observations made here shall provide a good overview of the influence of the design parameters on the finger behaviour, providing guidance to the choice of finger parameters given the range of motion and forces required for a new application. The setups used can be seen in Fig. \ref{fig:exp_setup} \rev{and Fig. \ref{fig:stiffness_setup}}.

\subsection{Single Finger Stiffness}
\label{sec:single_finger_stiffness}
To measure the stiffness of a single finger, the edge of the \textit{Schunk} gripper module is moved downwards onto the contact surface of the finger, resulting in a normal force on the contact surface, as seen in Fig. \ref{fig:stiffness_setup}. The resulting force is measured via the F/T sensor seen in Fig. \ref{fig:robot_left}. To characterize the stiffness, movements in the $y$ and $z$ direction within the elastic range are applied, usually $10-15$ mm. \rev{The stiffness is identified with a least-squares fit for a linear stiffness model.}

\subsubsection{Stiffness in $y$}
To evaluate the stiffness of the experimental setup (finger fixture, robotic systems, robot pedestal), a solid finger is printed with 6 top layers, 6 bottom layers, 3 walls and a gyroid infill with a density of $20\%$. Its stiffness in $y$ is identified as $8.20$ N/mm, a factor of 4 higher than the compliant fingers. 

\begin{figure}[t]
\begin{center}
    \includegraphics[width=0.5\textwidth]{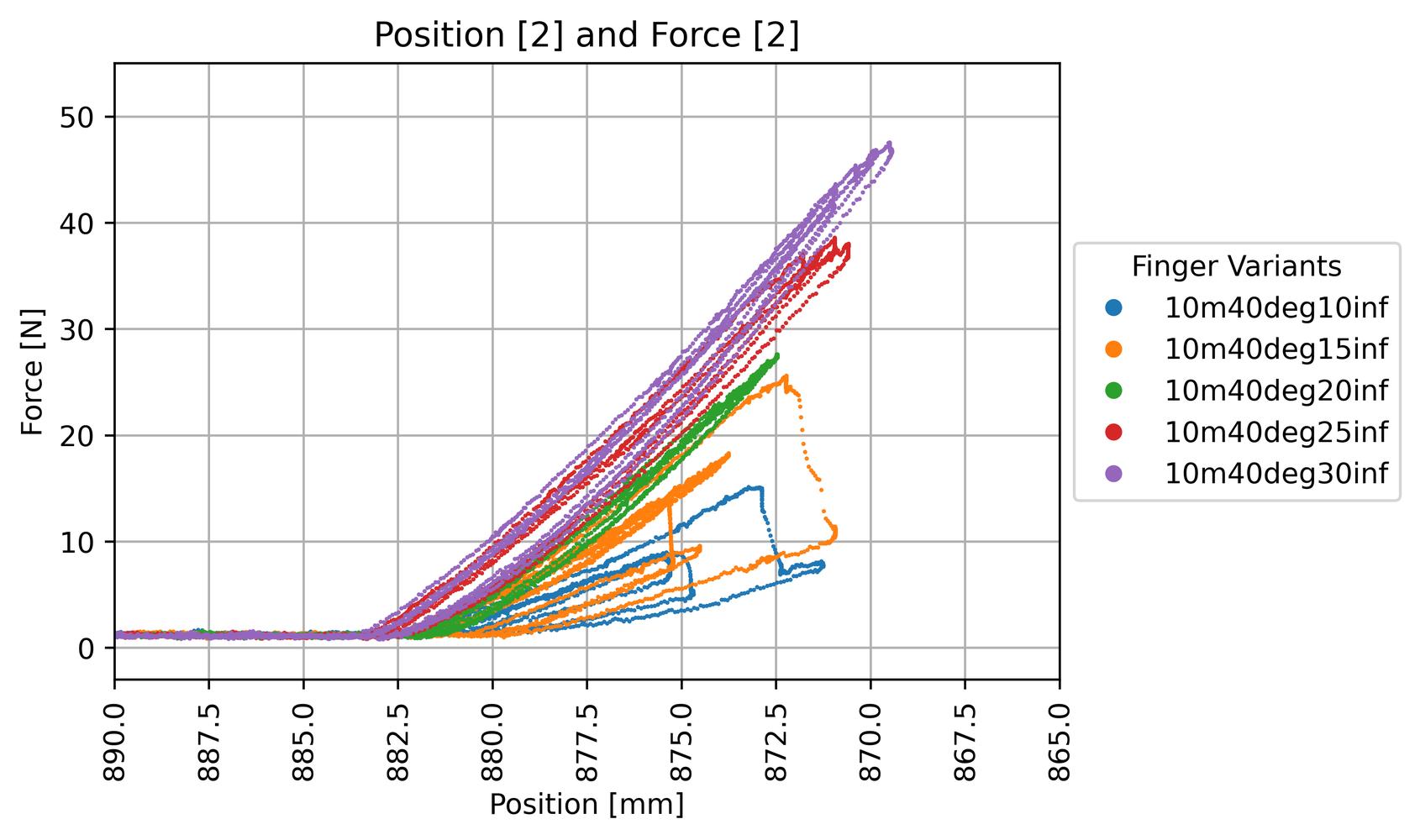}
    \end{center}
    \caption{The stiffness for varying infill density values, from 10, 15, 20, 25 and 30 \%,  using a  40° infill direction is demonstrated. Strong hysteresis is especially noticeable for 10 and 15 \% infill direction}
    \vspace{-0.5cm}
    \label{fig:stiffness_40deg}
    \end{figure}   

\begin{figure}[t]
\begin{center}
    \includegraphics[width=\columnwidth]{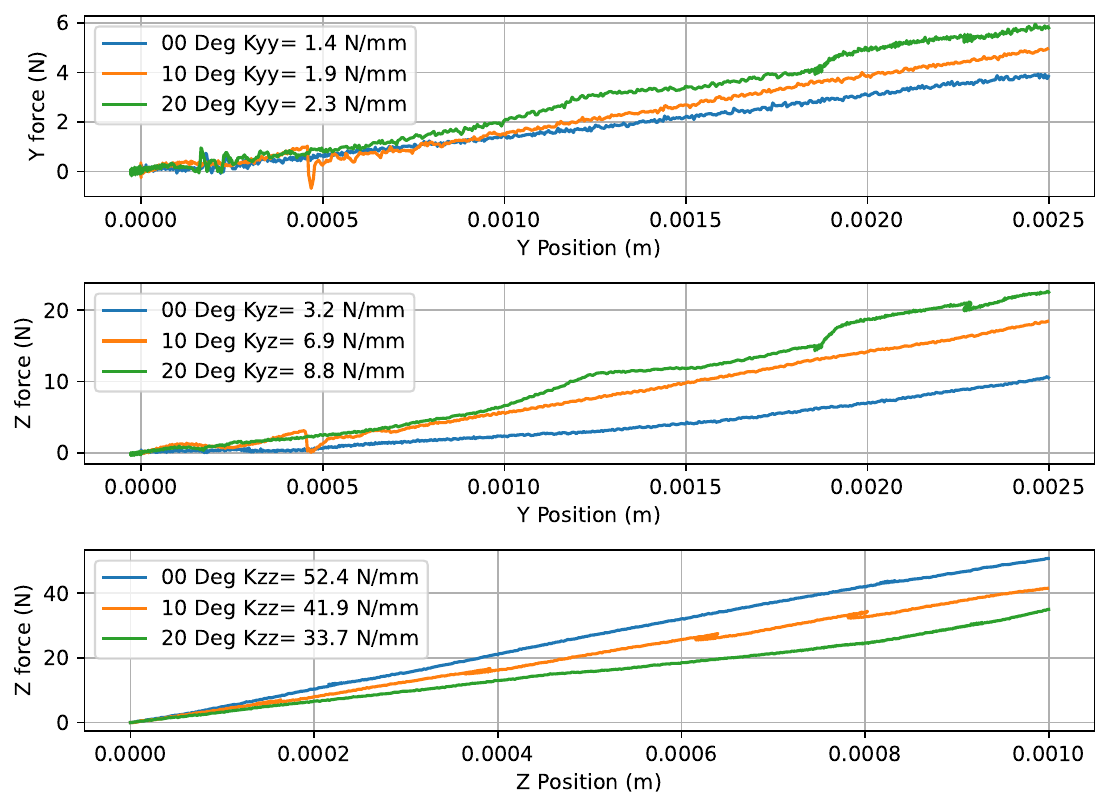}
    \end{center}
    \caption{The directional stiffness for infill directions of 0, 10 and 20$\degree$ with 10\% infill density.}
    \vspace{-0.5cm}
    \label{fig:stiffness_10deg}
    \end{figure}

 Fig. \ref{fig:stiffness_40deg} shows the stiffness of $40 \degree$ infill direction over various infill density values ($10 \%$ to $30 \%$). Here, hysteresis is strongly noticeable on lower infill densities due to plastic deformations of the finger, i.e. buckling, which can be seen in Fig. \ref{fig:stiffness_40deg}. An increase of stiffness with infill density is shown, where the maximum value is reached at $40 \degree$ infill direction and $30\%$ infill density, the maximum values tested. 

\rev{Further tests of linear stiffness in $y$ for other materials and infill densities} are listed in Tab. \ref{tab:fingers_single_stiffness}. Both $10\%$ and $15\%$ infill density are insufficient values for an infill direction of $40 \degree$ resulting in buckling, leading to low stiffness values, as seen in Tab. \ref{tab:fingers_single_stiffness}. Additionally, finger stiffness is more sensitive to infill density at lower infill directions, as seen in Tab. \ref{tab:fingers_single_stiffness}. 
As seen in Tab. \ref{tab:fingers_single_stiffness}, for the same parameters the PETG fingers achieve a lower stiffness compared to the PLA+ fingers, by roughly a factor of $2$. The same trend in increasing stiffness over infill density is seen. 

\subsubsection{Directional stiffness}
\rev{The remaining translational stiffnesses are measured on the setup of Fig. \ref{fig:robot_left} as
\begin{equation}
    \begin{bmatrix} F_x \\ F_y \\ F_z \end{bmatrix} = \begin{bmatrix} K_{xx}  & - & - \\ - & K_{yy} & K_{zy} \\ - & -K_{zy} & K_{zz} \end{bmatrix} \begin{bmatrix} \delta_x \\ \delta_y \\ \delta_z \end{bmatrix} \label{eq:stiff}
\end{equation}
where $F_\cdot$ and $\delta_\cdot$ are the forces and displacements in the coordinate system shown on Fig. \ref{fig:robot_zoom} and $K_{ab}$ the stiffness coupling a displacement in $a$ to a force in $b$. Stiffnesses $K_{xz}$ and $K_{xy}$ are assumed to be zero, with the only off-diagonal coupling within the grip plane. $K_{xx}$ is measured at $2.9$ N/mm for a $10\%$, $0\degree$ PLA+ finger, measured with the fingertip fixed. The stiffnesses $K_{yy}$, $K_{yz}$ and $K_{zz}$ are affected by the infill direction as can be seen in Fig. \ref{fig:stiffness_10deg}, where $K_{yz}\in[1.4,2.3]$ and $K_{zz}\in[33.7, 52.4]$ N/mm. 

This gives a stiffness ratio between assembly direction and transverse of $>13.7$, realizing a large difference in stiffness between directions. Additionally, the presence of $K_{yz}$ indicates remote center of compliance (RCC) effects \cite{huang1998, li2019c}. Diagonalizing the $K$ of \eqref{eq:stiff} yields the principle axis of the rendered stiffness \cite{huang1998}, which takes a direction of $3.6$, $9.5$ and $14.6\degree$ from $z$ for $0$, $10$, and $20\degree$ infill, respectively. This shows that the center of the RCC varies with the rib angles, where the lower infill directions provide an RCC farther from the gripper location.
}


\subsubsection{Viscoelastic effects}
\rev{ Velocity-dependent effects were checked by loading a $4$ mm displacement in the $y$ direction with a variety of velocities, ranging from $2$ to $15$ mm/s. A simple viscoelastic model was fit as $F_y = K_{xx}\delta_x + B_{xx}\dot{\delta}_x$, where $B_{xx}$ is the viscous term.  Fitting the terms on a $10\%$, $0\degree$ PLA+ finger found $K_{xx}=1.45$ N/mm and $B_{xx}=0.055$ Ns/mm, indicating at high speeds, e.g. $>100$ mm/s, the viscous terms will exceed $5$ N. }

\begin{table}[h!]
  \centering
  \caption{The results of the single finger stiffness experiments, stiffness in the y-direction is measured in $N/mm$. The stiffness of a solid PLA+ finger is $8.2 N/mm$.}
  \label{tab:fingers_single_stiffness}

  \begin{adjustbox}{width=\textwidth/2}

\begin{tabular}{|l||c|c|c|c|c|}
\hline
Infill    & $10\%$ infill  &  $15\%$ infill  & $20\%$ infill & $25\%$ infill & $30\%$ infill  \\
Dir. &   &    &  &  &   \\
$[\degree]$ & $[N/mm]$ & $[N/mm]$ & $[N/mm]$ & $[N/mm]$ & $[N/mm]$  \\

\hline
\hline

0  & PETG: 0.50   & - &  PETG: 1.133 & PLA+: 2.20 & PETG: 1.60   \\
    &   PLA+ : 1.2 &    &   PLA+: 1.95 &    &  PLA+: 3.00  \\
\hline
10 & PLA+: 1.05 & - & PLA+: 2.10 & - & PLA+: 3.00 \\
\hline
20 & PLA+: 1.533 & - & PLA+: 2.60 & - & PLA+: 3.2 \\
\hline
30 & PLA+: 1.667 & PLA+: 3.00 & PLA+: 3.00 & PLA+: 3.20 & PLA+: 3.70 \\
\hline
40 & PLA+: 0.85 & PLA+: 1.22 & PLA+: 3.4 & PLA+: 3.6 & PLA+: 3.84 \\
\hline

\hline

\end{tabular}

  \end{adjustbox}
\vspace{-0.5cm}
\end{table}

\subsection{Ultimate Strength and Maximum Deflection}
\label{sec:max_force}
In the following, the ultimate strength for each finger is determined and listed in Tab. \ref{tab:fingers_single_force}, under which the component fails, either by breaking of the outer walls or the inner ribs or by buckling, usually of the outer walls. For this, the the tool is moved in the z-direction until a component failure occurs. The maximum deflection until component failure occurs is also listed in Tab \ref{tab:fingers_single_maxdeflection}. 
Tab. \ref{tab:fingers_single_stiffness} and Tab. \ref{tab:fingers_single_maxdeflection} display a trend that the maximum achievable deflection reduces with an increasing infill direction. The PETG fingers, which have a comparably low stiffness compared to the PLA+ fingers, achieve a comparable maximum deflection but lower maximum force due to their stiffness.
Tab. \ref{tab:fingers_single_maxdeflection} shows that increasing infill direction results in a lower maximum tolerable force. However, increasing the infill density results in a higher maximum tolerable force and as Tab. \ref{tab:fingers_single_maxdeflection} demonstrates, also increases maximum deflection. The maximum tolerable force and the largest deflection occur for 0° infill direction with $30\%$ infill density. 

Additionally, buckling was noticed in the single finger stiffness experiments earlier, as mentioned in Sec.  \ref{sec:single_finger_stiffness}, where the threshold of the unstable behaviour for $40 \degree$ at $20 \%$ infill density is demonstrated compared to higher infill density values.

\begin{table}[h!]
  \centering
  \caption{The maximum tolerable force for various finger design parameters is demonstrated until mechanical failure occurs}
  \label{tab:fingers_single_force}

  \begin{adjustbox}{width=0.98\textwidth/2}

\begin{tabular}{|l||c|c|c|c|c|}
\hline
Infill  & 10\% infill  &  15\% infill  & 20\% infill & 25\% infill & 30\% infill   \\
Direction &  &    &  &  &    \\
$[\degree]$ & $[N]$ & $[N]$ & $[N]$ & $[N]$ & $[N]$  \\

\hline
\hline

0  & PETG: 12.50  & - &  PETG: 26.50 & PLA+: 56.50 & PETG: 46.00 \\
  &  PLA+ : 26.00  &  &   PLA+: 40.50 &  &  PLA+: 72.50  \\
\hline
10 & PLA+: 26.25 & - & PLA+: 45.00 & - & PLA+: 69.00 \\
\hline
20 & PLA+: 20.50 & - & PLA+: 41.50 & - & PLA+: 55.50\\
\hline
30 & PLA+: 16.50 & PLA+: 23.00 & PLA+: 34.00 & PLA+: 39.00 & PLA+: 51.00\\
\hline
40 & PLA+: 15.75 & PLA+: 17.00 & PLA+: 32.00 & PLA+: 38.50 & PLA+: 43.00 \\
\hline

\hline

\end{tabular}

  \end{adjustbox}
 \vspace{-0.5cm}
\end{table}

\begin{table}[h!]
  \centering
  \caption{The maximum deflection is shown as the deflection when mechanical failure occurs}
  \label{tab:fingers_single_maxdeflection}

  \begin{adjustbox}{width=0.95\textwidth/2}

\begin{tabular}{|l||c|c|c|c|c|}
\hline
Infill  & 10\% infill  &  15\% infill  & 20\% infill & 25\% infill & 30\% infill   \\
Direction &  &    &  &  &    \\
$[\degree]$ & $mm$ & $mm$ & $mm$ & $mm$& $mm$  \\

\hline
\hline

0  & PETG: 24.00  & - &  PETG: 22.50 & PLA+: 28.00 & PETG: 27.00  \\
  &  PLA+ : 26.75  &  &   PLA+: 27.00 & PLA+: 28.00 &  PLA+: 28.5  \\
\hline
10 & PLA+: 21.75 & - & PLA+: 23.00 & - & PLA+: 26.00 \\
\hline
20 & PLA+: 17.00 & - & PLA+: 19.00 & - & PLA+: 21.00\\
\hline
30 & PLA+: 13.75 & PLA+: 13.50 & PLA+: 15.50 & PLA+: 15.50 & PLA+: 16.50\\
\hline
40 & PLA+: 10.50 & PLA+: 10.00 & PLA+: 11.50 & PLA+: 13.50 & PLA+: 14.00 \\
\hline

\hline

\end{tabular}

  \end{adjustbox}
\vspace{-0.5cm}
\end{table}

\subsection{FEA}

\rev{To iterate a design quickly, an efficient evaluation method is needed. Printing the proposed fingers takes roughly an hour, which can support quick iteration. However, especially in an early design phase, an Finite Elements Analysis (FEA) can provide faster evaluation of mechanical properties. This section performs FEA for the fingers, and compares to the values of the single finger stiffness in Tab. \ref{tab:fingers_single_stiffness}.  
For the FEA, a static stress study is done in Autodesk Fusion 360, where it is assumed that the material (Sunlu PLA+ and Verbatim PETG) is isotropic. This is inaccurate due to the working principle of the extruder, where material is dispensed as a line with which the part is built layer-wise, resulting in a non-isotropic, directional material behaviour \cite{Xiao.2021}, but a needed assumption for feasibility. Additionally, approximations in the mesh process may impact the results due to the thin walls.}

\rev{Verbatim provides a technical data sheet with material properties determined according to various norms \cite{verbatim_PETG}, with a tensile strength of: $50$MPa, the Young's modulus with: $2050$MPa, and the density of $1.27$g/cm$^3$. 
Since Sunlu does not provide any mechanical properties, an assumption has to be made, where an already existing material (Profile: "Printed PolyTerra PLA Plastic") from an external database \cite{Graves2021} is used to model the variants, as the results of pre-tests have shown a high level of conformity with the actual material behavior. The values given are: density: $1,14$g/cm$^3$, the Young's modulus with: $1900$MPa, the Poisson's ratio with: $0.36$, the yield strength with: $20.04$MPa and the ultimate tensile strength with: $20.9$MPa}

\rev{
The force is set, according to the stiffness in Tab. \ref{tab:fingers_single_stiffness}, so that the displacement should be $\approx 1$ mm, and a probe measures the displacement, identifying the linear stiffness as $F_{yy} = K_{yy}\delta_y$. The resulting displacement, stiffness and the relative deviation from the stiffness values in Tab. \ref{tab:fingers_single_stiffness} of the analyzed fingers are listed in Tab. \ref{tab:FEA}.  \rev{Overall, the relative deviation, as seen in Tab. \ref{tab:FEA} is comparably low, deviating between 3\% and 26\%, demonstrating a robust performance of the FEA. The stronger deviations are discussed next.

Especially the results for $40\degree$ infill direction and 10\% infill density are noteworthy.
As the first entry for $40\degree$ infill direction and 10\% infill shows, there is a considerable discrepancy between the calculated stiffness value and the measured value from the experiments. This is because of a component failure of the finger at the experiment, where the deformation is no longer elastic but plastic due to a buckling of the structure. This is not considered in the FEA explaining the discrepancy of both values. To verify the FEA of the model, another approach is needed, demonstrated in the second entry.   
The second entry for $40\degree$ infill direction and 10\% infill is calculated with a force value by linearly extrapolating the two prior stiffness values at $20\degree$ infill direction and $3\degree$ with 10\% infill by
$K_{yy} = ax + b$, where $x$ is the infill direction, $a$ the linear effect, and $b$ the offset. With $K_{yy} = 1.667$ N/mm for $x=30\degree$ and $K_{yy}= 1.533$ for $x=20\degree$, $a=0.0134$ N/mm$\degree$. The extrapolated stiffness at $40\degree$ infill direction is then estimated at $1.801$ N/mm. This value is now used to derive the applied force for the FEA, resulting in a displacement of $\approx 0.97$ mm and a stiffness of $1.86$ N/mm resulting in a deviation of $\approx 3\%$ compared to the extrapolated value giving a high level of consistency. However, it becomes clear that the FEA does not correspond well to the actual measured values where instabilities occur in the real-life experiment as these are not taken into account or do not occur in the FEA, hence the deviation of $\approx 54 \%$.

As Tab. \ref{tab:FEA} demonstrates, the results for the PETG fingers deviate by $\approx 52\%$. Since the offset is relatively constant over the variants, it could be assumed that the deviation does not come from the model or the model's properties itself, but rather from the mechanical attributes provided by Verbatim or the missing Poisson's ratio. As discussed before, the term "tensile strength at yield" could be ambiguous and other important mechanical properties are missing. 

The used material shows a high level of conformity with the real-life experiments, as long as no instabilities have been present, resulting in deviations between $3\%$ and $26\%$. The relatively high deviation of $54\%$ is discussed before, where structural instabilities are accountable for the high deviation. While a good approximation, if the FEM setup requires substantial time (e.g. to re-construct the finger geometry otherwise determined by the slicer), the print-and-test iteration cycle may be more efficient.
}

\begin{table}[h!]
  \centering
  \caption{FEA results, where the displacement over the applied force is used to determine the respective stiffness which is then compared to the stiffness value in Tab. \ref{tab:fingers_single_stiffness}. $**$ denotes extrapolated force value.}
  \label{tab:FEA}

  \begin{adjustbox}{width=\textwidth/2}

\begin{tabular}{|l||c|c|c|c|c|}
\hline
Finger Type   & Displacement  & Applied Force & Stiffness  &  Relative Deviation \\
$ $ & $[mm]$ & $[N]$ & $[N/mm]$  & $[\%]$  \\

\hline
\hline

0° Infill Direction  &  PLA+: 0.94 &  PLA+: 1.2 &  PLA+: 1.28 &  PLA+: 6\\
10\% Infill Density & PETG: 0.41 & PETG: 0.5 & PETG: 1.22 & PETG: 59 \\
\hline

20° Infill Direction & PLA+: 1.07 & PLA+: 2 & PLA+: 1.43 & PLA+: 7 \\
10\% Infill Density &  & & & \\

\hline
30° Infill Direction & PLA+: 1.06 & PLA+: 1.733 & PLA+: 1.64 & PLA+: 6 \\
10\% Infill Density &  & & & \\
\hline
40° Infill Direction & PLA+: 0.46 & PLA+: 0.85 & PLA+: 1.85 & PLA+: 54 \\
10\% Infill Density &  & & & \\
\hline
40° Infill Direction & &  &  & \\ 
10\% Infill Density $**$ & PLA+: 0.97 & PLA+: 1.801& PLA+: 1.86 & PLA+: 3 \\
\hline

0° Infill Direction &  PLA+: 0.74 &  PLA+: 1.95 &  PLA+: 2.64 &  PLA+: 26 \\
20\% Infill Density & PETG: 0.54 & PETG: 1.133 & PETG: 2.1 & PETG: 46\\

\hline
20° Infill Direction & PLA+: 1.03 & PLA+: 2.6 & PLA+: 2.52 & PLA+: 3 \\ 
20\% Infill Density &  & & & \\
\hline
30° Infill Direction & PLA+: 1.19 & PLA+: 3.0 & PLA+: 2.52 & PLA+: 19 \\ 
20\% Infill Density &  & & & \\
\hline
40° Infill Direction & PLA+: 1.15 & PLA+: 3.4 & PLA+: 2.96 & PLA+: 15 \\
20\% Infill Density &  & & & \\
\hline
0° Infill Direction & PLA+: 0.82 & PLA+: 3.0 & PLA+: 3.66 & PLA+: 18\\
30\% Infill Density & PETG: 0.48 & PETG: 1.6 & PETG: 3.33 & PETG: 52\\
\hline
10° Infill Direction & PLA+: 0.91 & PLA+: 3 & PLA+: 3.3 & PLA+: 9\\
30\% Infill Density &  & & & \\
\hline
20° Infill Direction & PLA+: 0.94 & PLA+: 3.2 & PLA+: 3.4 & PLA+: 6\\
30\% Infill Density &  & & & \\
\hline
30° Infill Direction & PLA+: 1.07 & PLA+: 3.7 & PLA+: 3.46 & PLA+: 7\\
30\% Infill Density &  & & & \\
\hline
40° Infill Direction & PLA+: 0.94 & PLA+: 3.84 & PLA+: 4.09 & PLA+: 6\\
30\% Infill Density &  & & & \\
\hline
\end{tabular}

  \end{adjustbox}

\end{table}
\vspace{-0.5 cm}

}

\section{Validation}
\label{sec:app}

\begin{figure}[h!]
\begin{center}
    \subfloat[Gear assembly]{\includegraphics[width=0.33\columnwidth,trim={8cm 0 9cm 0},clip]{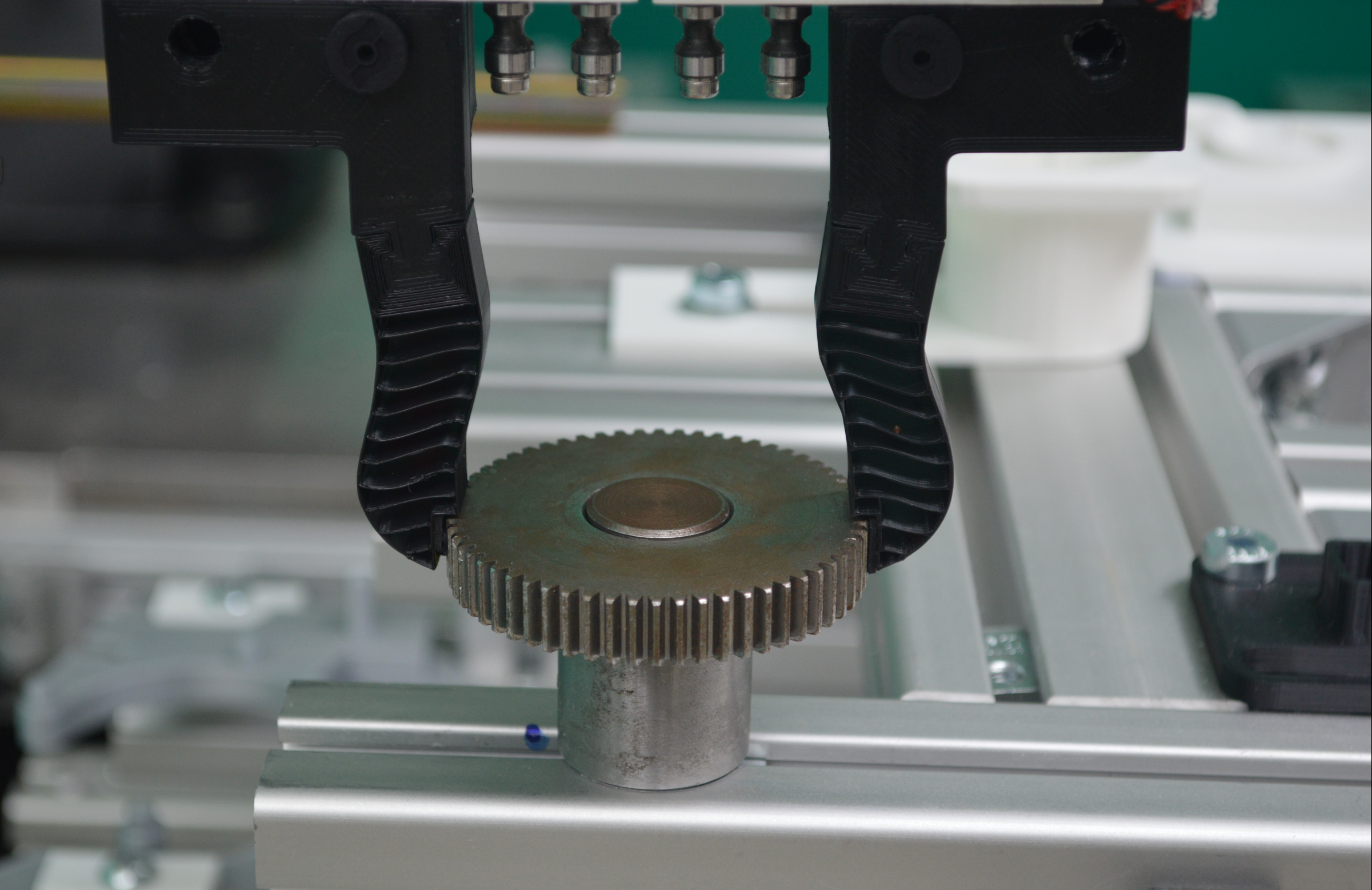}}
    \subfloat[Shuko plug]{\includegraphics[width=0.322\columnwidth, trim={11cm 2cm 8cm 0},clip]{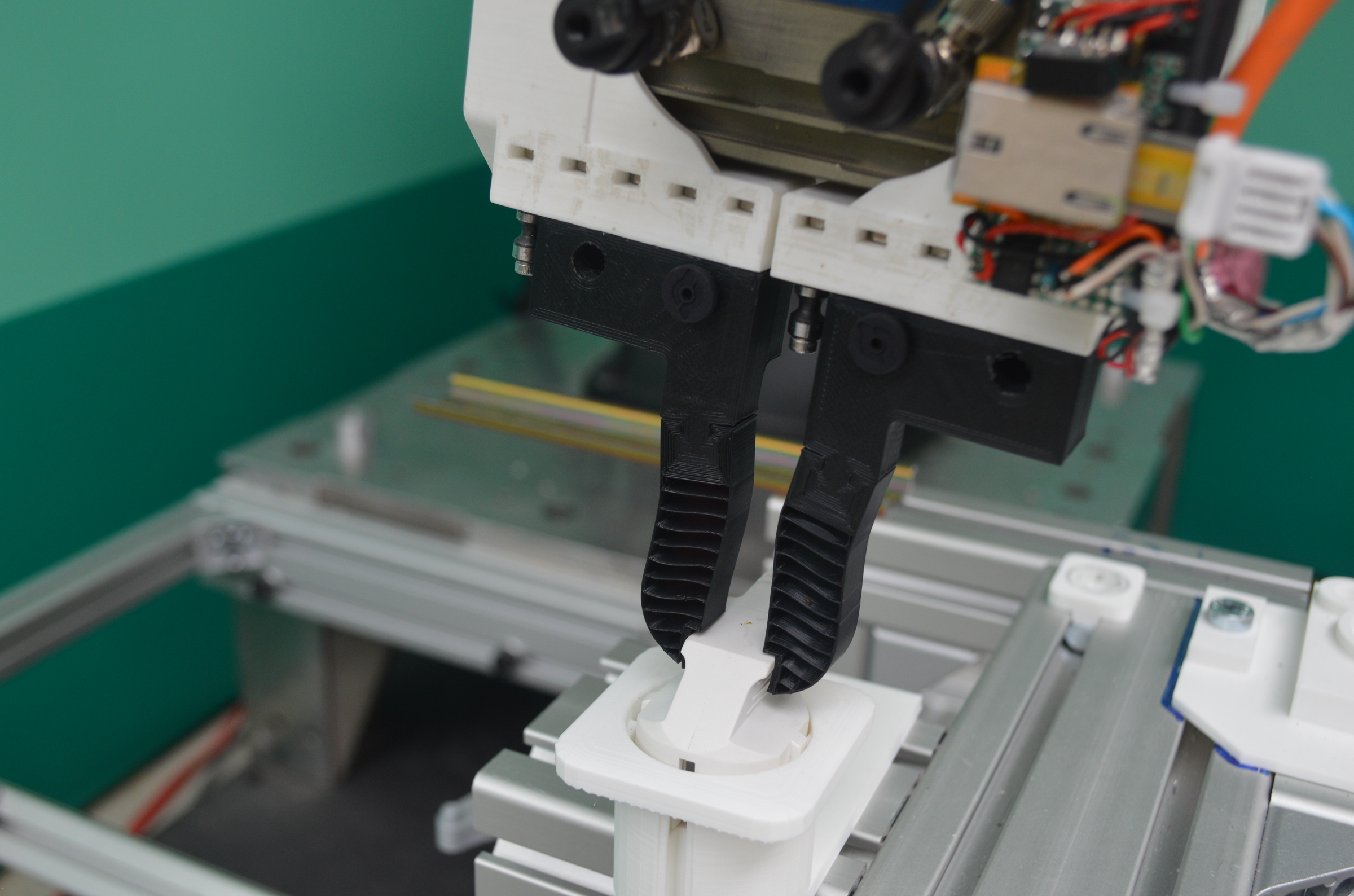}}
    \subfloat[PCB plug]{\includegraphics[width=0.32\columnwidth, trim={12cm 0 7cm 2cm},clip]{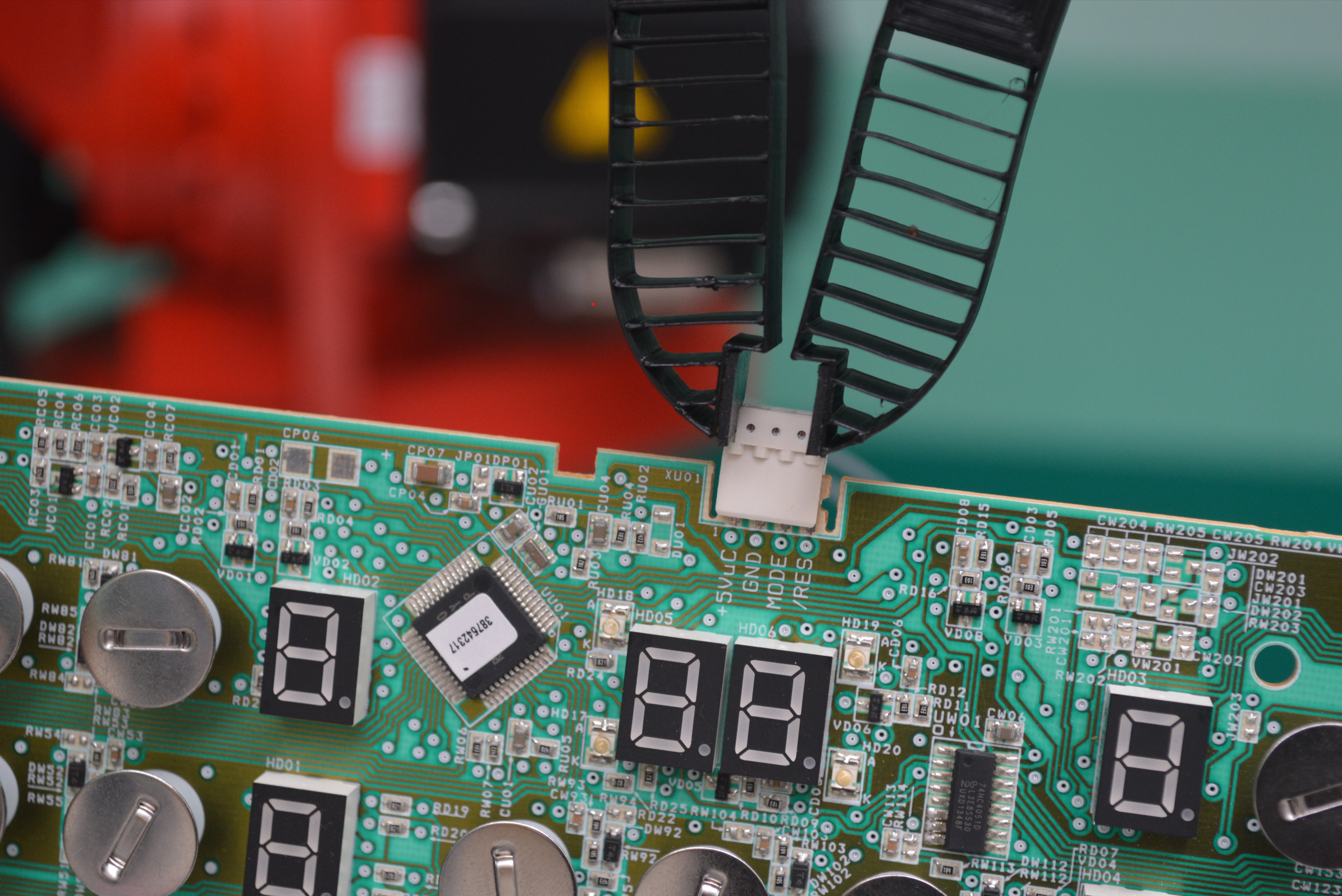}}
\end{center}
\caption{Additional assembly scenarios, see attached video. \label{fig:addl_scenarios}}
\vspace{-0.5 cm}
\end{figure}

This section validates the finger's performance in assembly applications. \rev{In addition to the assembly process shown in Fig. \ref{fig:working principle}, various objects shown in Fig. \ref{fig:addl_scenarios} can be assembled by the fingers.} The remainder of this section iteratively validates the fingers for the assembly of a plug into socket. The setup used is the same as in the previous experiments, demonstrated in Fig. \ref{fig:robot_zoom}, implementing the working principle presented in Fig. \ref{fig:working principle}.  

\begin{figure}[t]
\begin{center}
    \includegraphics[width=0.5\textwidth]{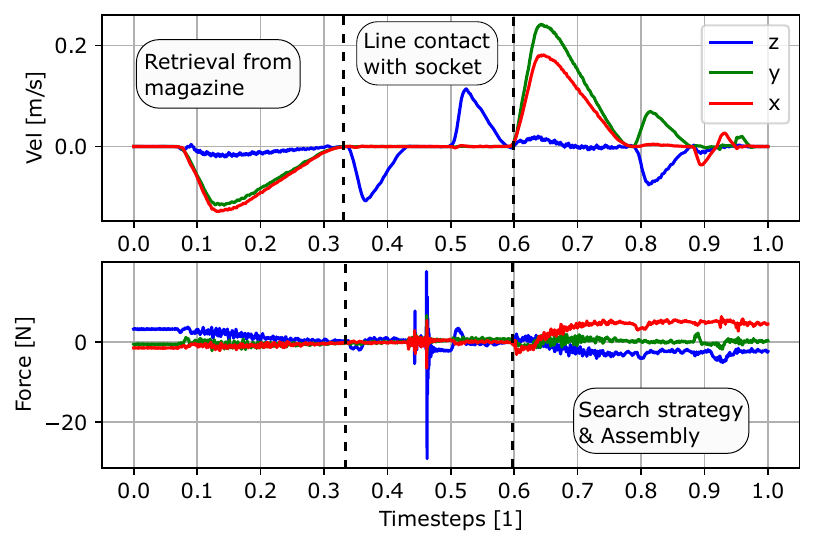}
    \end{center}
    \caption{Velocity and corresponding force measurements in the $x-$, $y-$ and $z-$direction for a $5 \%$ speed override value during assembly.}
    \vspace{-0.5cm}
    \label{fig:5ovr}
    \end{figure}

The goal is to successfully pick the plug from a magazine and assemble into a socket with and without various misalignment values. 
The program is intended for high-speed assembly, with tool speed values of $250 mm/s$, as seen in Fig. \ref{fig:5ovr}, up to over $1.3 m/s$\rev{.}
In Fig. \ref{fig:5ovr}, the force peak is attributable to the forces caused during the line contact while moving in z-direction, visualized in Fig. \ref{subfig:key-c} to fully assemble the parts. 


The assembly and grasping process is summarized as follows and can be seen in Fig. \ref{fig:robot_zoom}. First, the cable is grabbed from a magazine which provides a repeatable starting point. The contact force of the fingers overcome the contact force of the spring when moving the gripper in a linear movement upwards, with which the cable is removed from the magazine. To compensate slippage during the first phase, afterwards the gripper could push the cable head slightly on the table with a linear movement downwards to ensure a contact with the upper contact surface of the finger. Now, in the second phase, assembly takes place, using the search strategy described earlier in Section \ref{subsec:strategies}. A video is available \rev{as supplementary material and} at \url{https://youtu.be/J7EGXtE54oY}. 



\subsection{Repeatability Experiments}
\label{sec:repeatability_magazine}

To test for repeatability, the assembly process is repeated 84 times with a fixed socket position, using the aforementioned UR program and manually resetting the plug in the magazine, out of which the assembly failed twice. The first failure occurred at attempt 30 and the second failure at attempt 84 which ultimately lead to a component failure of the fingers. This concludes a roughly $97.6\%$ success rate. 
Assumed causes are either slight slippage in the grip coming from the adhesive tape, or the kinematics of the robot. Additionally, the table on which the robot is mounted is not fixed but on wheels, which could add another level of instability, impairing the robustness. 

\subsection{Robustness Experiments}
\label{sec:robustness_exp}

In the next experiments, the robustness over variation in socket position is tested, to clarify the impact of the design parameters on the robustness and tolerable range.  To control the misalignment, instead of a fixed waypoint for the socket position, a variable waypoint is programmed which can be changed in each iteration. For the initial test, the boundaries of compensable misalignment of the plug to the socket are determined in x- and y-direction in $0.5$ mm steps. A finger with $0 \degree$ infill direction, $10\%$ infill and a $10 \degree$ mount are used. A successful assembly is repeated five times to assure repeatability. If five assemblies in a row are successful, another 0.5 mm is added to the misalignment and the sequence of five trials starts again. This is repeated until the maximum compensable misalignment is met and the assembly fails for the first time. This is done to test the limits both in the  x- and y-direction. With this setup, it can be shown that with a 100\% speed value of the program and the used search algorithm this compliant finger design is capable of tolerating a misalignment in a range of $7.5$ mm in y-direction and $7$ mm in the x-direction. 
To further compare the tolerance windows with varying designs, the limits of the first run are tested with varying infill densities and infill directions. The results are listed in Table \ref{tab:erg_robustness}. 

\begin{table}[ht]
  \centering
  \vspace{0.2cm}
  \caption{Results Robustness Experiment, $m$ denotes mount, meaning which mount configuration is used (either 10° and 20°), $i$ denotes the infill density in percentage and $id$ is an abbreviation for the infill direction in deg. \rev{f denotes failure} } 
  \label{tab:erg_robustness}

     \begin{adjustbox}{width=0.95\textwidth/2}

    \begin{tabular}{|c|c|c|c|c|c|c|c|c|c|c|}
 & x & y & x & y & y & y & y & y & y & y\\
 & 10° m & 10° m  & 20° m & 20° m & 10° m & 10° m & 10° m & 10° m & 10° m & 10° m  \\

Variant & 10\% i & 10\% i & 10\% i & 10 \% i & 20\% i & 30\% i & 10\% i & 20\% i & 30\% i  & 10\% i \\

 & 0° id & 0° id  & 0° id & 0° id & 0° id & 0° id & 0° id & 0° id & 0° id & 10° id   \\
 & PLA+ & PLA+ & PLA+ & PLA+ & PLA+ & PLA+ & PETG & PETG & PETG & PLA+\\

\hline
\hline

range [mm] & 7 & 5.5 & 4.5 & 5.5 & 5 & 4.5 & 7.5 & 6 & 5.5 & 5.5  \\

\hline 

    \end{tabular}

  \end{adjustbox}

  \vspace{5pt}
    \begin{adjustbox}{width=0.95\textwidth/2}
    \begin{tabular}{|c|c|c|c|c|c|c|c|c|c|c|c|c|}
 & y & y & y & y & y & y & y & y & y & y & y \\
 & 10° m & 10° m & 10° m & 10° m & 10° m & 10° m & 10° m & 10° m & 10° m & 10° m & 10° m \\

Variant& 10 \% i & 10 \% i& 15 \% i & 20 \% i & 25 \% i& 30 \% i & 10 \% i & 15 \% i & 20 \% i & 25 \% i& 30 \% i  \\

 & 20° id & 30° id & 30° id & 30° id & 30° id & 30° id & 40° id & 40° id & 40° id & 40° id & 40° id  \\
 & PLA+ & PLA+ & PLA+ & PLA+ & PLA+ & PLA+ & PLA+ & PLA+ & PLA+ & PLA+ & PLA+ \\

\hline
\hline

range [mm] & 5 & 4 & 5.5 & 5.5 & 5.5 & 6 & f & f & f & 2 & 5.5  \\

\hline

    \end{tabular}    
 \end{adjustbox}

\end{table}

It is important to note that regarding the compensable range in x-direction, the compensation is attributable to the free rotation of the cable head inside the grip about the y-axis, corresponding to the coordinate system visualized in Fig. \ref{subfig:key-a}, and should be treated as a positive side-effect of the finger's design, which can be described as an unforeseen DoF. 
The main focus of the finger's design is to allow a compliance in the y-direction due to the ribs, which is why the experiment only shows general feasibility of a compensation in the x-direction for the $10 \degree$ and $20 \degree$ mounts but does not compare the tolerance in x-direction for every finger, as seen in Tab. \ref{tab:erg_robustness}.

\subsection{Discussion}
As Tab. \ref{tab:erg_robustness} shows, with a $10 \degree$ mount the tolerable misalignment-range is slightly larger than with a $20 \degree$ mount. During the tests for $20 \degree$ infill direction and 10 \% infill density at a misalignment of $5$ mm early signs of buckling are noticed. At the next increment of the infill direction, this is noticed already at $4$ mm and plastic deformations at the connections of the ribs to the outer wall appear at $4.5$ mm. $30 \degree$ infill direction with 10 \% infill initially stands out due to a comparably large compensable range of $\approx 5.5 - 6.5$ mm. However, beginning with $+4$ mm misalignment, some slight buckling can be observed building up to slight plastic deformation at the connection of the ribs to the outer wall with the next misalignment increments, which is why this variant is not considered ideal.  
The extreme value of $40 \degree$ direction shows to be difficult to test. At a comparably low misalignment value of $2$ mm, component failure already occurs for $10\%$ infill density resulting in a non-feasible combination for any assembly tasks. This is also noticeable for $15\%$ infill density where buckling and component failure occurred at $3$ mm and $3.5$ mm. Because the initial start-value is set too high, the part is already permanently damaged, resulting in a failed assembly at $-1.5$ and $-2$ mm. At $20 \%$ infill there is strong bulging noticeable at $2$ mm and buckling at $2.5$ mm. $-2$ mm proves to be compensable, however, bulging is noticeable here, too. With $25 \%$ infill strong deformation is noticed at $3.5 mm$. $4 mm$ is also successful, however, some plastic deformation occurs, which is why the experiment is stopped here to prevent any further damage. The last increment for the infill density at $40 \degree$ infill direction proves to be the most stable one. Some strong deformation is observed at $3 mm$ but without plastic deformation. At $-3$ mm the cable head strongly clips into the plug, which is why no further tests are done for this variant to prevent any further damage. This is attributable to an excessive vertical stiffness of the finger, where compliance is still present, with a potentially too high contact force profile which could damage the electrical components. Thus, this variant should not be used to assemble delicate parts.  

To compensate misalignment parallel to the moving direction of the gripper's jaws, structured compliance in the base-y-direction is desired. However, unintentional DoFs, as a rotation about the base-y-axis or the base-x-axis resulting in a change of pose of the grasped part resulting from contact forces, are possible. 


Regarding the PETG fingers, $0\degree$ infill direction and $10\%$ infill density proves to achieve the biggest tolerance range of all combinations, tolerating $\approx7.5$ mm misalignment. However, this combination is not suitable for any assembly tasks because the cable head slips easily inside the grip. This is traced to a very low gripping force from the fingers due to the infill settings. Increasing the infill density by $10\%$ already results in a better grip, while achieving a tolerance range of $\approx6$ mm. Another $10\%$ show similar results, the compensation of $+3.5$ mm misalignment cannot be repeated robustly. Using PETG comes with the benefit of a higher flexibility compared to PLA+, which results in a lower risk of plastic deformation during handling. 
The tolerable range can be defined as $\approx5.5$ mm while providing a stable grasp on the gripped cable head.

\section{Conclusion and Future Work}

\label{sec:discussion}
To the authors' best knowledge, this work has proposed a first use of a finray-effect gripper for structured compliance in assembly. That is, compared with previous works using the finray-principle which focus on a stable grasp with varying object surface geometry, this design realizes directionally-dependent stiffness on the gripped part. This is used to robustly and repeatedly compensate misalignment in the range of up to $7.5$ mm in high-speed assembly tasks. Additionally, the objective, as defined before in Sec. \ref{sec:intro}, of achieving a comparable success time as in \cite{park2013intuitive} is reached and exceeded, as the assembly time from  first contact is $\approx 1.2$ seconds. Hence, feasibility of the passive compliant fingers to compensate misalignment in high-speed tasks without additional sensing is proven. 
With an increasing infill density and increasing infill direction, the stiffness of the finger can be increased, as shown in Tab. \ref{tab:fingers_single_stiffness}. However, an increasing infill direction results in an earlier structural failure of the finger, in both maximum force and deflection. 
For an optimal finger design, the finger stiffness, the ultimate strength, the maximum deflection, the gripping stability and the compensable range have to be taken into consideration. A variant with a too high stiffness, e.g. variants with a $30\%$ infill density, especially with an increasing infill direction could damage the assembly parts. A too low stiffness, e.g. PETG with $10 \%$ infill density would not be able to lift and transport the cable robustly and maintain a stable grip when external forces occur. Choosing a $40 \degree$ infill direction results in component failure due to plastic deformation especially at lower infill density values.
Most of the variants listed in Tab. \ref{tab:erg_robustness} achieve a tolerable range of $\approx 5.5$ mm, $30 \%$ infill density and $0 \degree$ infill direction achieves the lowest, with $\approx 4.5$ mm. PETG shows the best results here, with a maximum range of $\approx 7.5$ mm for the non-applicable $10 \%$ infill variant. Thus, the higher rib angle PETG variants are recommended in this case. 


Future work will focus on establishing additional attempts to design the fingers by using FEA or by analytically determining the mechanical properties and to achieve a better intuition of how the design parameters influence the final stiffness of the structure. 

Using fused deposition modeling as an additive manufacturing process comes with its own limitations, as the direction in which the part is built up has to be considered. Certain structures need an optimal orientation to the print bed to be successfully manufactured, as overhangs or otherwise unsupported structures could fail without support. Using alternative manufacturing processes could allow one to create ribs in varying directions which could introduce multi-directional structured compliance into the finger. Additionally, other material could be used which could achieve higher contact forces and would be less sensitive to wear and fatigue.  

\bibliographystyle{IEEEtran}
\bibliography{lib2.bib} 

\end{document}